\documentclass[conference]{IEEEtran}
\IEEEoverridecommandlockouts
\usepackage{cite}
\usepackage{amsmath,amssymb,amsfonts}
\usepackage{graphicx}
\usepackage{textcomp}
\usepackage{xcolor}
\usepackage{multirow}

\usepackage{caption}
\usepackage{subcaption}

\usepackage{booktabs}
\usepackage{algorithm}  
\usepackage{algpseudocode}

\usepackage{amsthm}

\def\BibTeX{{\rm B\kern-.05em{\sc i\kern-.025em b}\kern-.08em
    T\kern-.1667em\lower.7ex\hbox{E}\kern-.125emX}}

\newtheoremstyle{ICDEdefinition} 
    {3pt}   
    {3pt}   
    {\normalfont} 
    {}      
    {\bfseries} 
    {.}     
    {.5em}  
    {#1\ #2.\ (#3)}

\theoremstyle{ICDEdefinition}
\newtheorem{definition}{Definition}

\begin{document}

\title{FedDW: Distilling Weights through Consistency Optimization in Heterogeneous Federated Learning\\
\thanks{*Corresponding Author}
}


\author{
    \IEEEauthorblockN{
        Jiayu Liu\textsuperscript{1},
        Yong Wang\textsuperscript{1}\textsuperscript{2}\textsuperscript{*},
        Nianbin Wang\textsuperscript{1},
        Jing Yang\textsuperscript{1},
        Xiaohui Tao\textsuperscript{3}
    }
    \IEEEauthorblockA{
        \textsuperscript{1}College of Computer Science and Technology, Harbin Engineering University, Harbin, China \\
        \textsuperscript{2}Modeling and Emulation in E-Government National Engineering Laboratory, Harbin Engineering University, Harbin, China \\
         \textsuperscript{3}School of Mathematics, Physics and Computing,  University of Southern Queensland, Toowoomba, Australia
    }
    \IEEEauthorblockA{
        \{jiayuliu, wangyongcs, wangnianbin, yangjing\}@hrbeu.edu.cn, xiaohui.tao@unisq.edu.au
    }
}

\maketitle

\begin{abstract}
Federated Learning (FL) is an innovative distributed machine learning paradigm that enables neural network training across devices without centralizing data. While this addresses issues of information sharing and data privacy, challenges arise from data heterogeneity across clients and increasing network scale, leading to impacts on model performance and training efficiency. Previous research shows that in IID environments, the parameter structure of the model is expected to adhere to certain specific consistency principles. Thus, identifying and regularizing these consistencies can mitigate issues from heterogeneous data. We found that both soft labels derived from knowledge distillation and the classifier head parameter matrix, when multiplied by their own transpose, capture the intrinsic relationships between data classes. These shared relationships suggest inherent consistency. Therefore, the work in this paper identifies the consistency between the two and leverages it to regulate training, underpinning our proposed FedDW framework. Experimental results show FedDW outperforms 10 state-of-the-art FL methods, improving accuracy by an average of 3\% in highly heterogeneous settings. Additionally, we provide a theoretical proof that FedDW offers higher efficiency, with the additional computational load from backpropagation being negligible. The code is available at https://github.com/liuvvvvv1/FedDW. 
\end{abstract}
\section{Introduction}
Federated Learning (FL) \cite{b1},\cite{b2},\cite{b3} has gained significant attention as a privacy-preserving machine learning (ML) paradigm that enables multiple clients to collaboratively train models without sharing their raw data. Traditional FL frameworks, such as FedAvg \cite{b4}, assume homogeneous data distribution and similar computational capabilities across clients. However, in real-world scenarios, data heterogeneity and varying computational resources among clients are inevitable, posing significant challenges to the efficiency and effectiveness of FL systems.

\begin{figure}
\centering
\includegraphics[width=\columnwidth]{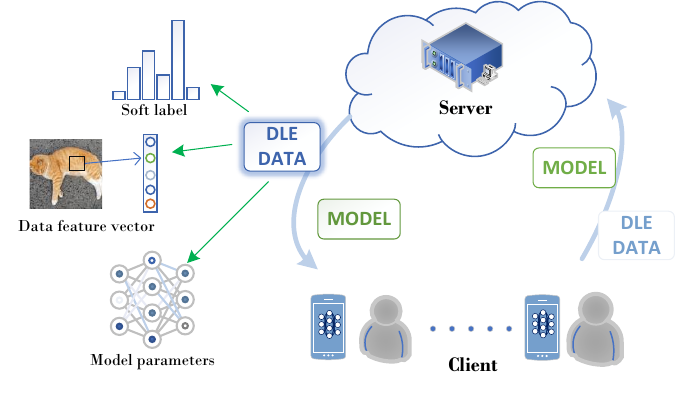}
\caption{Schematic diagram of DLE data transmission in the federated system and three common deep encryption data.}
\label{fig:1}
\end{figure}

Data heterogeneity \cite{b5},\cite{b6},\cite{b7} often referred to as non-IID (Non-Independent and Identically Distributed) data, can degrade the performance of federated models and increase the divergence of local updates from different clients \cite{b8}. Because of the singularity of data distribution, each client has a different emphasis on data distribution. Therefore, while each client has the same objective function and model structure, the value of the optimal solution is often different. Just like in the long-tail problem of data \cite{b9}, \cite{b10}, if a certain class of data accounts for a large proportion of the data set, the model tends to be more inclined toward this class of data rather than learning a reasonable model weight. The main method to solve heterogeneous federated learning is information sharing between clients. One client must know the data information of other clients (i.e. global information) to train a model with global generalization, but this process will violate confidentiality agreements. To resolve this conflict, we introduce deep learning encrypted data (\textbf{DLE data}), which is defined in detail as follows.

\begin{definition}[\textbf{DLE Data}]
The data \(e\) is generated during the neural network training process and encapsulates the characteristics of the training dataset. However, under current technological conditions, it cannot be effectively decoded to reconstruct the actual data or derive specific statistical results from the dataset.
\end{definition}

So far, there are three common types of DLE data, as shown in Figure~\ref{fig:1}. One is the data feature vector \cite{b11}, \cite{b12}, \cite{b13}, \cite{b14}. It uses a local neural network as a feature extractor to map high-dimensional image or text data into a low-dimensional feature space and use it as the feature vector of the image. The process of mapping high-dimensional data into a low-dimensional feature vector is irreversible. It not only retains the core features of the data, but also reduces the transmission cost during sharing. This is the most common DLE data. The second is soft labels \cite{b15},\cite{b16},\cite{b17}. This type of data applies knowledge distillation \cite{b18} techniques to FL. This type of DLE data is more concerned with the relationship between different categories of data rather than the data itself. The third is the neural network model \cite{b19}, \cite{b20}. After the neural network model is trained on a dataset, the parameters inside the model will reflect some of the characteristics and information of the dataset. The client shares it with other clients and, due to the black box characteristics and unexplainability of the neural network itself, it can become an excellent encryption method. The disadvantage of deep learning unexplainability turns into an advantage for encryption.

\begin{figure}
\centering
\includegraphics[width=\columnwidth]{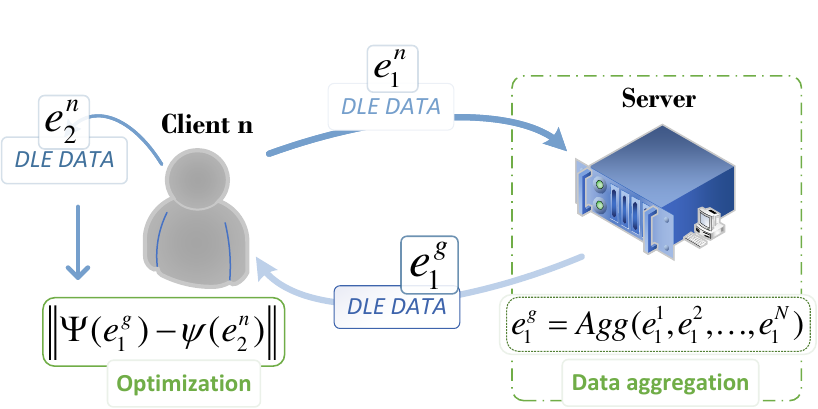}
\caption{The client uses DLE data for regular optimization, generating two types: \(e_1\) for global aggregation to capture global information, and \(e_2\) to guide optimization toward global generalization by aligning with \(e_1\). \(e_1\) must be chosen to ensure it retains generalization after aggregation. In particular, \(e_1\) and \(e_2\) may be equal.}
\label{fig:2}
\end{figure}

Here, we propose a consistency optimization paradigm to address the heterogeneity issue using DLE data. The detailed definition is as follows.


\begin{definition}[DLE Consistency Optimization]
Under IID data distribution, find two types of DLE data: \(e_1\) and \(e_2\) with consistent relationship. For the \(n\)-th client, it involves optimizing the consistency objective: \(\left\| \Psi(e_1^{g}) - \psi(e_2^{n}) \right\|,\)
where \(e_1^{g} = Agg(e_1^1, e_1^2, \dots, e_1^N)\). Here, \( \Psi \) and \( \psi \) denote the Mapping Functions, and \( Agg(\cdot) \) represents the aggregation operation performed on the server.
\end{definition}


The entire DLE consistency optimization process is shown in Figure~\ref{fig:2}. This paper identifies the consistency relationship between soft labels and classification layer parameters and, based on this foundation, proposes FedDW. FedDW regularizes the parameters of the local model's classification layer using these global soft labels, ensuring that under non-IID data distributions, client models retain the parameter distribution characteristics of IID environments. In summary, the main contributions of the paper are the following:
\begin{itemize}
\item This paper first proposes and defines the concept of DLE data, providing three common examples of DLE data.
\item This paper presents the discovery of a consistency relationship between soft labels and the classification layer parameter matrix multiplied by their own transpose, further verified through experiments.
\item The paper proposes a concise and novel FedDW framework, proving both experimentally and theoretically that FedDW has better performance and training efficiency.
\end{itemize}

\section{Background and Related Work}
After \cite{b21} demonstrated that transmitting gradients could also leak local information, FedAvg\cite{b4} became the most effective FL method. It is simple to implement and has very competitive performance. However, FedAvg does not perform well with heterogeneous local data. As the degree of heterogeneity increases, the performance of the FedAvg method declines rapidly. To address this issue, FedProto \cite{b14} clients use server aggregated average feature vectors to regularize local training. FedGH \cite{b11} uses these feature vectors as training data to train the classifier part of the model. FedTGP\cite{b12}, leveraging the idea of contrastive learning \cite{b22},\cite{b23},\cite{b24},\cite{b25}, aligns and enhances the performance of heterogeneous data and models from different clients in FL through shared trainable global prototypes and adaptive boundary contrastive learning. FedKTL\cite{b26} achieves effective local transmission of pre-trained knowledge to the client through image vector pairs generated by the server.
\begin{figure*}
\centering
\includegraphics[width=\textwidth]{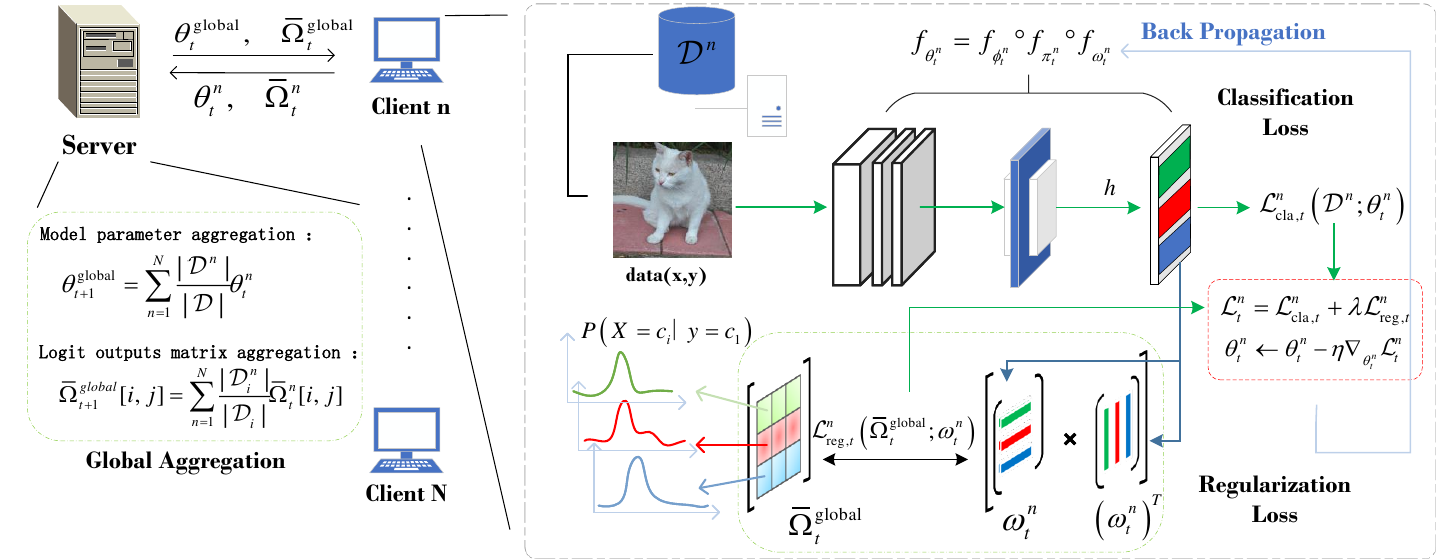}
\caption{The entire training process of FedDW.}
\label{fig:3}
\end{figure*}

FedProx\cite{b6} and FedDyn\cite{b14} are classic FL frameworks that use neural network parameters as globally shared information. The objective function of the local model introduces a proximal term, preventing local updates from straying too far from the initial global model. Moon\cite{b20} ensures that the features extracted by the local model are closer to the features obtained by the global model transmitted by the server while simultaneously being further away from the features extracted by the local model from the previous training round. FedHEAL\cite{b27} maintains the variance distance between the local model and the global model during local training on the client. 

PerAda\cite{b15}, MH-pFLID\cite{b16}, Fedistill\cite{b17}, DaFKD\cite{b28}, and FedAUX\cite{b29} all use soft labels as global shared information. Their core idea is to aggregate the average soft labels generated by all clients and use them for regular local training through different methods.

There are other FL methods that do not use any DLE data but can still mitigate the problem of data heterogeneity, such as FedRep\cite{b30}, FedALA\cite{b31}, FedPer\cite{b32}, and FedRAP\cite{b33}. These methods often fall under Personalized Federated Learning(pFL)\cite{b34}. pFL does not produce a single model but rather personalizes a model for each user based on their data distribution. However, deep learning training requires large amounts of data, and pFL trains or fine-tunes most of the network structure using only locally stored data, which is insufficient. Thus, as the name pFL itself suggests, it cannot provide a generalized model on the server side. This limitation restricts its range of applications.

Using DLE data to mitigate data heterogeneity introduces additional computational costs. None of the aforementioned methods attempt to alleviate the extra computational burden. With the increasing size of models in today’s training landscape, especially with the advent of large models\cite{b35},\cite{b36},\cite{b37}, the additional computational cost of regularizing data heterogeneity is something that absolutely cannot be ignored. In this regard, FedDW offers a significant advantage.

\section{Methodology}
\subsection{Problem Statement}
We assume that there are \( N \) clients, denoted \( P_1, \ldots, P_N \). Client \( P_n \) has a local dataset \( \mathcal{D}^n \). Our goal is to learn a Deep Neural Network model \( \theta \) over the dataset \( \mathcal{D} \triangleq \bigcup_{n \in [N]} \mathcal{D}^n \) with the help of a central server, while the raw data are not exchanged. The objective is to solve
\begin{equation}
\arg\min_\theta\mathcal{L}(\theta)=\sum_{i=1}^N\frac{|\mathcal{D}^n|}{|\mathcal{D}|}\mathcal{L}^n(\theta), \label{eq:1}
\end{equation}
where \( \mathcal{L}^n(\theta) = \mathbb{E}_{(x,y) \sim \mathcal{D}^n} [\ell(\theta; (x, y))] \) is the local model loss of \( P_n \).
For the convenience of analysis, we divide the DNN model into feature extractors \(f_{\phi}\) and classifiers \(f_{c}\), as inspired by \cite{b25}. We divide the classifier into \textbf{Mapping Layer}  \(f_{\pi}\) and \textbf{Classification Layer} \(f_{\omega}\). The classification layer is the network parameters of the last layer of the classifier, while the rest is the mapping layer. So the entire model is \(f_{\theta} = f_{\phi} \circ f_{\pi} \circ f_{\omega} \).
\subsection{Distribution Matching}
The core idea of FedDW is inspired by the field of contrastive learning. In contrastive learning, data \(\mathbf{X}\) is mapped into a feature vector \(h \in \mathbb{R}^k \)
. The similarity relationship between data \(\mathbf{x}_{i}\) and data \(\mathbf{x}_{j}\) can be expressed as the Cosine Similarity of their corresponding feature vectors \(\mathbf{h}_{i}\) and \(\mathbf{h}_{j}\)(i.e.,\( \text{Sim}(\mathbf{h}_{i}, \mathbf{h}_{j}) =\mathbf{h}_{i}^\top \mathbf{h}_{j} / \| \mathbf{h}_{i} \| \| \mathbf{h}_{j} \| \)). We remove the bias term of the classification layer. When the classification layer uses the feature vector \(\mathbf{h}\) for calculation, the calculation formula is as follows
\begin{equation}
\omega\mathbf{h} = \left[ \omega_{c_1}, \omega_{c_2}, \ldots, \omega_{c_{|\mathcal{C}|}} \right]^\top \mathbf{h} = \left[ \hat{P}_{c_1}, \hat{P}_{c_2}, \ldots,\hat{P}_{c_{|\mathcal{C}|}} \right]^\top
, \label{eq:2}
\end{equation}
where \({c}_i\) represents the i-th class, and \( \mathcal{C} = \bigcup_{i=1}^{|\mathcal{C}|} {c}_i\)
represents all data classes of all clients. The value generated by the vector of the i-th row of the weight matrix \( \omega \in \mathbb{R}^{|\mathbf{C}| \times k}\) of the classification layer corresponds to the probability of the \({c_i}\)-class. From the perspective of contrastive learning, \(\omega_{c_i}\) belongs to the feature space of \({c_i}\)-class. It obtains the probability of the corresponding class by calculating the inner product of vector with \(\mathbf{h}\). Contrastive learning calculates the Cosine Similarity between vectors, and the difference between the two is the modulus division operation. Under IID distribution, the modulus division operation here can be omitted, that is, the inner product can also replace the Cosine Similarity to express the correlation. The reasons are outlined as follows.

\begin{figure*}
\centering
\includegraphics[width=\textwidth]{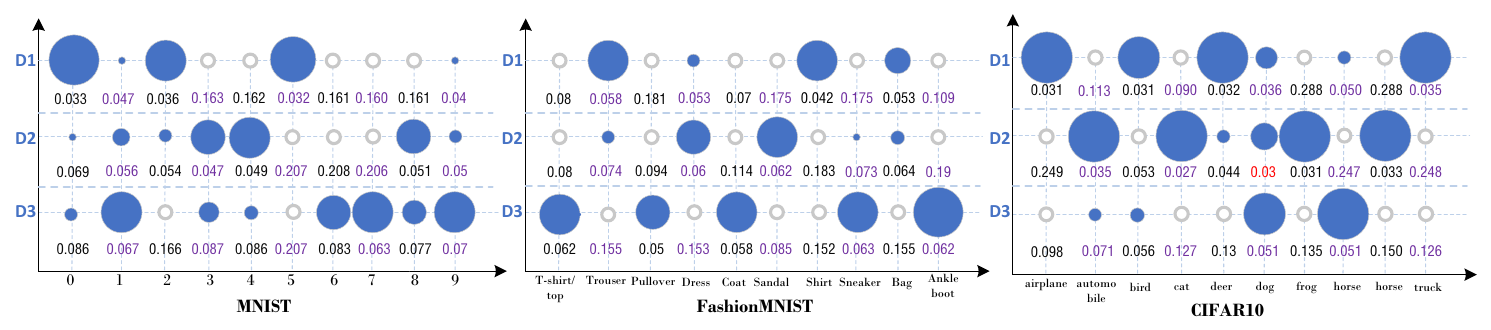}
\caption{We test each dataset 100 times. Due to space constraints, only three different data (D1, D2, D3) are randomly selected here to display the modulus values. The size of the blue circle represents the proportion of the data in this class. The larger the circle, the more data there is in this class. The white circle means there is no data in this class. The number under the circle represents the relative size of the weight vector modulus of the class corresponding to this class, while the calculation formula of other classes is \( \frac{\| \omega_{c_k} \|_2}{\sum_{i=1}^{|\mathbf{C}|} \| \omega_{c_i} \|_2}\). The red number represents the counterexample shown. In 100 experiments on three data sets, we only found 3 counterexamples.}
\label{fig:4}
\end{figure*}

Empirical results show that the modulus of the neural network classification layer weight \( \|\omega_{c_i}\| \) correlates with the proportion of \( c_i \)-class data. We designed an experiment to validate this empirical conclusion. Using MNIST, FashionMNIST, and CIFAR-10 datasets under zero bias, we varied class proportions and observed consistent trends (Figure~\ref{fig:4}). Surprisingly, \( \|\omega_{c_i}\| \) is \textbf{negatively correlated} with the proportion of class \( c_i \)-class data. Higher proportions lead to smaller weight moduli. This counter-intuitive result highlights the impact of class distribution on weight norms. Notably, results were not averaged across multiple experiments on a single data distribution. For IID data, weight moduli should be uniform, then the modulus division operation does not affect the relative size of the probability value corresponding to each class. Thus, the inner product of feature vectors approximates similarity. We define two properties in feature space: modulus, reflecting class proportions, and inner product, indicating inter-class similarity. In IID datasets, uniform moduli ensure the inner product reflects similarity effectively.

\begin{figure}
\centering
\includegraphics[width=\columnwidth]{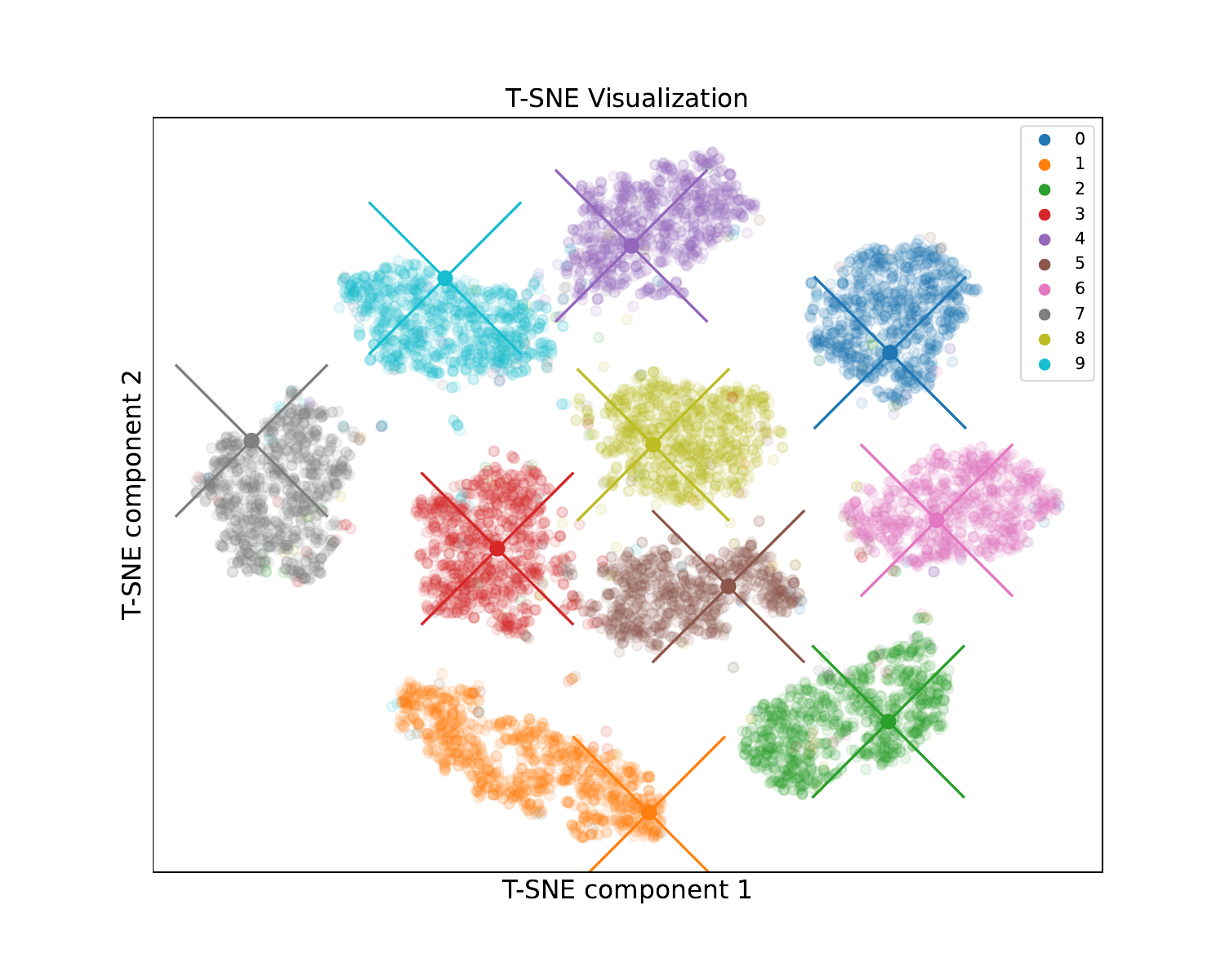}
\caption{We use "X" to represent the weight vector of each class of data in the classification layer, and the midpoint of "X" represents the specific position of the weight vector in the visualization space. We can see that each weight vector belongs to the cluster of the corresponding class. Note that, before visualization, we need to perform \textbf{Vector Unitization}.}
\label{fig:5}
\end{figure}

If we input the \(\omega_{c_i}\) into the classification layer network, then the probability value of \({c_i}\)-class should be the highest, and the values of other classes reflect its similarity with other classes. We use the weights of the classification layer to construct a correlation matrix, multiply it by its own transpose, calculated as follows:
\begin{equation}
\omega \cdot \omega^\top = 
\begin{bmatrix}
    \omega_{c_1} \\
    \omega_{c_2} \\
    \vdots \\
    \omega_{c_{|\mathcal{C}|}}
\end{bmatrix}
\begin{bmatrix}
  \omega_{c_1}^\top & \omega_{c_2}^\top & \cdots & \omega_{c_{|\mathcal{C}|}}^\top
\end{bmatrix}
. \label{eq:4}
\end{equation}
where \(\omega\omega^\top[i,j] \) means the similarity between the \({c_i}\)-th and \({c_j}\)-th data. Hereinafter, the matrix \(\omega\omega^\top\)  will be referred to as \textbf{CR (Class Relation) matrix}. We observe that the CR matrix and soft labels convey the same meaning, representing inter-class similarity. In IID settings, there is consistency between the two. Then under non-IID conditions, clients can use global soft labels to regularize the CR matrix, ensuring that the classification layer retains the parameter distribution structure of the IID scenario during training. This regularization further induces adjustments in other parameters, helping to mitigate issues caused by data heterogeneity.

Finally, we validate the above hypothesis through a visualization experiment: \(\omega_{c_i}\) belongs to the clustering of \({c_i}\)-class data in the feature space constructed by neural networks. After training a simple neural network consisting of two convolutional layers and three linear layers on the MNIST dataset, we use t-SNE to visualize the feature vectors of the data and the weight vectors of each class in the classification layer. The visualization effect is shown in Figure~\ref{fig:5}. The weight vector of the classification layer corresponding to each class is visualized within the cluster of its respective class. This confirms the validity of our hypothesis.

\subsection{The FedDW Framework}
\subsubsection{Global Aggregation}

During the t round of communication, the server will perform a weighted average of soft labels collected from the client according to the class label to obtain the average soft labels of each class. Then, the average soft labels of all classes are stacked in the order of class labels to construct a global average soft labels matrix \( \overline{\Omega}_t^\text{global}\in\mathbb{R}^{|\mathcal{C}|\times|\mathcal{C}|}\). In the following paper, this matrix is referred to as \textbf{SL matrix}. Simultaneously, in order to reduce the communication bandwidth, each client constructs a local average SL matrix according to the above method. Then, it is sent to the server together with the number of classes in the local dataset. The calculation formula of the client's local average SL matrix \( \overline{\Omega}_t^n\in\mathbb{R}^{|\mathcal{C}|\times|\mathcal{C}|}\) is as follows
\begin{equation}
\overline{\Omega}_{t}^{n}[i,j]=\frac{1}{|\mathcal{D}_{i}^{n}|}\sum_{\begin{array}{c}(x,y)\in \mathcal{D}^{n},\\y=c_{i}\end{array}}\widehat{p}_{t,j}^{n},\label{eq:4}
\end{equation}

\begin{equation}
\hat{P}_t^n = \sigma \left( f_{\phi_t^n} \circ f_{\tau_t^n} \circ f_{w_t^n}(x) \right),\label{eq:5}
\end{equation}

\begin{equation}
\hat{P}_t^n = \left[ \hat{P}_{t, 1}^n, \hat{P}_{t, 2}^n, \ldots, \hat{P}_{t, |\mathcal{C}|}^n \right]^\top.\label{eq:6}
\end{equation}
where \(\sigma\) represents the softmax operation.

Note that uploading the average SL matrix may leak user privacy. There are many ways to solve this problem. For example, the server might share a public dataset \cite{b15} or use differential privacy technology \cite{b38}. This paper does not go into details here. Based on the information uploaded by all clients, the calculation formula of the global SL matrix is 
\begin{equation}
\overline{\Omega}_{t+1}^{global}[i,j]=\sum_{n=1}^{N}\frac{|\mathcal{D}_{i}^{n}|}{|\mathcal{D}_{i}|}\overline{\Omega}_{t}^{n}[i,j].\label{eq:7}
\end{equation}
where \( \mathcal{D}_{i}^{n}\) represents the number of data of the i-th class in the dataset of the n-th client. Of course, in addition to obtaining the global SL matrix (like Fedavg) the server also performs a weighted average calculation of the model parameters as shown below:
\begin{equation}
\theta^{\text{global}}_{t+1} = \sum_{n=1}^{N} \frac{|\mathcal{D}^n|}{|\mathcal{D}|} \theta^n_t.\label{eq:8}
\end{equation}
\subsubsection{Client Local Training}

During the FL training process, each client needs to use its own data to independently train the local network parameters in parallel. During the t round of communication, the client body \( P_n \) receives the global model \(\theta^{\text{global}}_{t}\) and the global SL matrix \( \overline{\Omega}_{t}^{\text{global}} \) sent by the server, then replaces the original local model with the global model for local training. The global SL matrix is used for regularization during the training process. Therefore, the loss function of local training is divided into two parts, with one being the loss formed by training the neural network parameters with the local data set. This paper uses the classification problem as the problem to be solved. We call this part of the loss the classification loss and record it as \( \mathcal{L}_{cla,t}^{n}\). The calculation formula is as follows: 
\begin{equation}
\mathcal{L}_{cla,t}^{n}=\mathbb{E}_{(x,y)\sim D^{n}}[\ell(\theta_{t}^{n};(x,y))]
,\label{eq:9}
\end{equation}
\begin{equation}
\ell(\theta_{t}^{n};(x,y))=-\sum_{i=1}^{|\mathcal{C}|}\log(\widehat{P}_{t,i}^{n})[y==c_{i}]
.\label{eq:10}
\end{equation}

The other part is the loss formed by the regularization training process, denoted as \( \mathcal{L}_{\text{reg},t}^{n}\). Here we will use the global SL matrix to regularize the training. According to the above analysis, we know that after removing the bias term of the last linear layer, when training on an IID distributed data set, the CR matrix and the global SL matrix have the same numerical meaning. Both \(SL[i,j]\) and \(CR[i,j]\) express the similarity between the \({c_i}\)-data and \({c_j}\)-data. Then, the two matrices should be numerically equal. We utilize the above equal properties to regularize the parameters under the non-IID data distribution. According to the definition of DLE Consistency Optimization in our introduction, \(e_1\) is the SL matrix and \(e_2\) is the CR matrix, which regularizes the parameters based on \(e_1\). \( \Psi \) represents an equivalence mapping and \( \psi \) refers to the matrix operation of multiplying by its own transpose, followed by softmax. We use Frobenius Distance as a measure between two matrices, and use this as the regularization term. This makes the last parameter of the neural network sequence contain the properties under the IID data distribution, thereby adjusting other network parameters to achieve the effect of alleviating data heterogeneity. The final regularization calculation formula is
\begin{equation}
\mathcal{L}_{reg,t}^{n}(\overline{\Omega}_{t}^{global},\omega_{t}^{n})=\frac{1}{|\mathcal{C}|\times|\mathcal{C}|}||\overline{\Omega}_{t}^{global}-\sigma(\omega_{t}^{n}(\omega_{t}^{n})^\top)||_{F}^{2}
. \label{eq:11}
\end{equation}
Here \(\sigma\) is the softmax operation on the rows of the matrix.
So in summary, we obtain the entire objective function for locally updating parameter at iteration t as
\begin{equation}
 \mathcal{L}_{t}^{n} := \mathcal{L}_{\text{cla},t}^{n}\left(\mathcal{D}^{n};\theta_{t}^{n}\right)
+
\mu\mathcal{L}_{\text{reg},t}^{n}\left(\overline{\Omega}_{t}^{\text{global}};{\omega}_{t}^{n}\right)
. \label{eq:12}
\end{equation}
where \(\mu\) is the hyperparameter for trading off classification term  and regularization term , 
The process of local updates can be expressed as follows
\begin{equation}
 \theta_{t}^{n} \leftarrow \theta_{t}^{n} - \eta \nabla_{\theta_{t+1}^{n}} \mathcal{L}_{t}^{n} \left(\mathcal{D}^{n}; \theta_{t}^{n}; \overline{\Omega}_{t}^{\text{global}} \right)
. \label{eq:13}
\end{equation}
where \(\eta\) is the learning rate. Finally, Algorithm~\ref{alg:1} presents the entire training process of FedDW Framework. The entire algorithm process is shown in Figure~\ref{fig:3}.

\section{Theoretical Analysis}
\subsection{Convergence Analysis}

To analyze the theoretical nature of FedDW, we introduce Ordered Pair \( \langle t, r \rangle \). \( t \) indicates the communication round and \( r \in \{ \frac{1}{2}, 1, \ldots, R \} \) is a local update iteration. \( \langle t, r \rangle \) is the \( r \)-th iteration in the \( t \)-th round, while \( \langle t, \frac{1}{2} \rangle \)indicates that at the beginning of the \( t \)-th round, the client has completed replacing its own model with the global model and successfully received $\overline\Omega_{t}^{\text{global}}$. 

Following the previous convergence analysis method\cite{b6},\cite{b11},\cite{b14},  we present the following three statistical assumptions regarding the gradient of model parameters.

\textbf{Assumption 1.} (Lipschitz Smoothness). The \( n \)-th local loss function is \( L_1 \)-Lipschitz smooth, so the gradient of the local loss function is \( L_1 \)-Lipschitz continuous, where \( L_1 > 0 \), \( \forall r_1, r_2 \in \left\{\frac{1}{2}, 1, \dots, R \right\} \):
\begin{equation}
\begin{aligned}
&\left\| \nabla \mathcal{L}_{\langle t, r_1 \rangle}^n \left( \mathcal{D}^n ; {\theta}_{\langle t, r_1 \rangle}^n \right) 
- \nabla \mathcal{L}_{\langle t, r_2 \rangle}^n \left( \mathcal{D}^n ; {\theta}_{\langle t, r_2 \rangle}^n \right) \right\|_2 \\
&\leq L_1 \left\| {\theta}_{\langle t, r_1 \rangle}^n - {\theta}_{\langle t, r_2 \rangle}^n \right\|_2.
\end{aligned}
\label{eq:14}
\end{equation}

\textbf{Assumption 2.} \textit{Unbiased Gradient and Bounded Variance.} The random gradient \( g_t^n = \nabla \mathcal{L}_t^n \left( \omega_t^n; \mathcal{B}_t^n \right) \) (\( \mathcal{B} \) is a batch of local data) of each client’s local model is unbiased:
\begin{equation}
\mathbb{E}_{\mathcal{B}_t^n \subseteq \mathcal{D}^n} \left[ g_t^n \right] = \nabla \mathcal{L}_t^n \left( \omega_t^n \right),
\label{eq:15}
\end{equation}
and the variance of random gradient \( g_t^n \) is bounded by:
\begin{equation}
\mathbb{E}_{\mathcal{B}_t^n \subseteq \mathcal{D}^n} \left[ \left\| \nabla \mathcal{L}_t^n \left( \omega_t^n; \mathcal{B}_t^n \right) - \nabla \mathcal{L}_t^n \left( \omega_k^t \right) \right\|_2^2 \right] \leq \sigma^2.
\label{eq:16}
\end{equation}

\textbf{Assumption 3.} The function \( \mathcal{L}_{\text{cla},t}^{n} \) is convex. Considering the complexity of the problem, the work in this paper only provides convergence analysis when the preservation function \( \mathcal{L}_{\text{cla},t}^{n} \) is convex. This is a strong condition, but we only apply it to theorem 2, to ensure that the expected loss of the aggregated model is always less than the weight sum of the expected losses of all clients before aggregation, i.e.,
\begin{equation}
\begin{aligned}\mathbb{E}[\mathcal{L}_{cla, \langle t+1,\frac{1}{2} \rangle}^{}]\leq\frac{1}{N}\sum_{n=1}^{N}\mathbb{E}[\mathcal{L}_{cla,\langle t,R \rangle}^{n}]\end{aligned}
.\label{eq:17}
\end{equation}

\textbf{Theorem 1} (The regularization term is bounded). \textit{The essence of regularity is the sum of the squares of the differences between the corresponding position elements of a matrix where two row elements are one. Easy to determine:}

\begin{equation}
0 < \mathcal{L}_{\text{reg}}^n < \frac{2}{|\mathcal{C}|}
.\label{eq:18}
\end{equation}

\textbf{Theorem 2.} (One-round deviation). \textit{Based on the above assumptions and Theorem 1, for an arbitrary client, after each communication round, we have:}
\begin{multline}
\mathbb{E}\left[\mathcal{L}_{\langle t+1, \frac{1}{2} \rangle} \right] \leq \mathbb{E}\left[\mathcal{L}_{\langle t, \frac{1}{2} \rangle} \right] 
- \left( \eta - \frac{L_1 \eta^2}{2} \right) \sum_{r=\frac{1}{2}}^{R} \left\| \nabla \mathcal{L}_{\langle t,r \rangle} \right\|_{2}^{2} \\
+ \frac{R L_{1} \eta^{2} \sigma^{2}}{2} + \frac{2 \mu}{|c|}
.\label{eq:19}
\end{multline}
\textit{The proof here refers to \cite{b14}, and Theorem 2 shows the deviation bound of the local objective function for a client after one communication round.}

\textbf{Theorem 3.} (Non-iid convergence rate of \texttt{FedDW}). \textit{Based on the above assumptions and theorems, and where \( \mathcal{L}^* \) refers to the local optimum. For a client, given any \( \epsilon > 0 \), after}
\begin{multline}
\frac{1}{T} \sum_{t=1}^{T} \sum_{r=\frac{1}{2}}^{R} \mathbb{E} \left[ \left\| \nabla C_{\langle t, r \rangle}^{n} \right\|_2^2 \right] 
\leq \frac{2 \left( \mathcal{L}_{\langle 1, \frac{1}{2} \rangle}^{n} - \mathcal{L}^{*} \right)}{T \left( 2 \eta -  L_1 \eta^2 \right)} \\
+ \frac{L_1 E \eta^2 \sigma^2 + 4 \mu}{(2 \eta -  L_1 \eta^2) |\mathcal{C}|}
\leq \epsilon
.\label{eq:20}
\end{multline}
when
\begin{multline}
T
\leq 
\frac{2 |\mathcal{C}| \left( \mathcal{L}_{\langle 1, \frac{1}{2} \rangle}^{n} - \mathcal{L}^{*} \right)}
{\epsilon |\mathcal{C}| \left( 2 \eta - L_1 \eta^2 \right) - |\mathcal{C}| L_1 \left( R \eta^2 \sigma^2 \right) - 4 \mu}
.\label{eq:21}
\end{multline}
and
\begin{multline}
\eta < \frac{\epsilon |\mathcal{C}| + \sqrt{\epsilon^2 |\mathcal{C}|^2 - 4 \mu L_1 |\mathcal{C}| (\epsilon - R \sigma^2)}}
{L_1 |\mathcal{C}| (\epsilon - R \sigma^2)}
.\label{eq:22}
\end{multline}
\textit{Therefore, each client’s local model can converge at the non-convex convergence rate \( \epsilon \sim \mathcal{O} \left( \frac{1}{T} \right) \). Theorem proof reference is in \cite{b11} - due to space limitations, detailed proof has been omitted.}

\subsection{Regularization Term Properties}
We analyze the properties of the regularization term itself. For the convenience of analysis, we transform term \( \sigma(\omega_{t}^{n}(\omega_{t}^{n})^\top) \) into matrix A, where A satisfies :
\begin{equation}
A = \left\{ A \in \mathbb{R}^{ |\mathcal{C}| \times  |\mathcal{C}|} \mid 0 \leq A_{ij} \leq 1, \sum_{j=1}^{n} A_{ij} = 1, \ \forall i, j \right\}.\label{eq:23}
\end{equation}
Meanwhile, we abbreviate \( \Omega_{t}^{\text{global}}\) as \( \Omega\), so we can convert the formula for  the regularization term \( \mathcal{L}_{reg} \) into \( \mathcal{L}_{reg}(A)=tr((\Omega-A^{\top}A)^{\top}(\Omega-A^{\top}A))\). Below, we provide the first derivative of \( \nabla \mathcal{L}_{reg}(A)\) and the Hessian matrix \( H(\mathcal{L}_{reg})\) corresponding to the second derivative:

\begin{equation}
\nabla \mathcal{L}_{reg}(A)=-2(\Omega-A^{\top}A)A,
\label{eq:24}
\end{equation}

\begin{equation}
H(\mathcal{L}_{reg})=2(AA^\top\otimes I)+2(I\otimes AA^\top)-2I\otimes(\Omega-A^\top A).
\label{eq:25}
\end{equation}
where \(I\) is the identity matrix and \( \otimes\) denotes the Kronecker product. It is found that \( \nabla \mathcal{L}_{reg}(A)\) only involves the parameter matrix of the last layer of the neural network, and has nothing to do with other parameters and the neural network structure. Computational cost generated by the regularization term does not increase with the expansion of the neural network model. The gradient generated by the regularization term is the same for any data. Then in a batch, the gradient only needs to be calculated once.

\( H(\mathcal{L}_{reg})\) has both positive and negative items, which indicates that \( H(\mathcal{L}_{reg})\) is not positive definite. Therefore,  \( \mathcal{L}_{reg}\) is not a convex function, although we did not require \( \mathcal{L}_{reg}\) to be convex in the convergence analysis above. But if it is a convex function, this will be more advantageous for model training. In practical training, we can construct convex functions by linearly approximating \( \mathcal{L}_{reg}\). 

We use \( A_0\) as a linearization reference point and perform a first-order Taylor expansion on \( A^TA\):

\begin{align}
A^{\top} A &\approx A_0^{\top} A_0 + A_0^{\top} (A - A_0) + (A - A_0)^{\top} A_0  \label{eq:26} \\
&\approx A_0^{\top} A + A^{\top} A_0 - A_0^{\top} A_0.
\label{eq:27}
\end{align}

We substitute it into the original equation, which has

\begin{align}
\mathcal{L}_{\text{reg}}(A) &= \text{tr}((\Omega - A^{\top} A)^{\top} (\Omega - A^{\top} A)) \label{eq:28} \\
&= \| \Omega - A^{\top} A \|_F^2 \label{eq:29} \\
&\approx \| \Omega - A_0^{\top} A - A^{\top} A_0 + A_0^{\top} A_0 \|_F^2.\label{eq:30} 
\end{align}

Since \( A_0\) is not updated as the initial matrix, the approximate function is a quadratic convex function of the matrix A. When using the iterative method to optimize the function, the reference point \( A_0\) can be updated at intervals of a certain number of iterations to keep approaching the optimal point. This method is similar to convex approximate optimization.

\begin{algorithm}
\caption{FedDW Framework}
\begin{algorithmic}[1]  
\Require The server and \( N \) clients; datasets of \( N \) clients \(\{D^n\}_{n=1}^N\); initial global model \(\theta_1^{\text{global}}\); initial global SL matrix \(\overline\Omega_1^{\text{global}}\); learning rate \(\eta\); trade-off factor \(\mu\).
\Ensure \(\theta_T^{\text{global}}\)

\State \textbf{Initialization:}
\State The server sends \(\theta_1^{\text{global}}\) and \(\overline\Omega_1^{\text{global}}\) to all clients.

\State \textbf{FedDW Communication:}
\For{\(t = 1, 2, \dots, T\)}
    \State \textbf{Clients:}
    \For{\textbf{each client} \(n = 1, 2, \dots, N\) \textbf{in parallel}}
        \State Initialize local model \(\theta_t^n \gets \theta_t^{\text{global}}\)
        \State Train local model \(\theta_t^n\) using Eq.~\eqref{eq:13}
        \State Compute local SL matrix \(\overline\Omega_t^n\) using Eq.~\eqref{eq:4}
        \State Upload \(\theta_t^n\) and \(\overline\Omega_t^n\) to the server
    \EndFor
    \State \textbf{Server:}
    \State Aggregate \(\{\theta_t^n\}_{n=1}^N\) to update \(\theta_{t+1}^{\text{global}}\) by Eq.~\eqref{eq:8}
    \State Aggregate \(\{\overline\Omega_t^n\}_{n=1}^N\) to update \(\overline\Omega_{t+1}^{\text{global}}\) by Eq.~\eqref{eq:7}
    \State Send \(\theta_{t+1}^{\text{global}}\) and \(\overline\Omega_{t+1}^{\text{global}}\) to all clients
\EndFor
\State \Return \(\theta_T^{\text{global}}\)

\end{algorithmic}
\label{alg:1}
\end{algorithm}

\section{Experiments}
\subsection{Experimental Setup}
We conduct a comprehensive comparison between FedDW and 10 popular FL frameworks including FedAvg\cite{b4}, FedUV\cite{b39}, MOON\cite{b20}, FedBABU\cite{b40},FedProto\cite{b14}, FedProx\cite{b6}, FedRep\cite{b30}, FedALA\cite{b31}, Per-FedAvg\cite{b41}, and FedPer\cite{b32}. In the CV domain, we select the MNIST dataset, the CIFAR-10 and CIFAR-100 datasets. In the NLP domain, we select the IMDB dataset. For different datasets, we apply different models. For the MNIST dataset, we use a simple two-layer CNN network as the feature extractor. For CIFAR-10, we add two max pooling layers after the CNN layers. For CIFAR-100, we use ResNet-18\cite{b42}. We use a bidirectional LSTM neural network as the feature extractor for the IMDB dataset. For the classifier part, all datasets use two fully connected layers as the mapping layers \( f_{\pi} \) and one fully connected layer as the classification layer \(f_{\omega} \). The size of vector passed to the classification layer for all models is 128.

We use PyTorch\cite{b43} to implement the FedDW framework and various comparison experiments. To ensure fairness in comparison, we use the Adam\cite{b44} optimizer with a learning rate of 0.001, and all other parameters are set to their PyTorch default values in all experiments. The hyperparameters of different frameworks are chosen to be as consistent with the original papers as possible. In the CV domain, the batch size is 128, while in NLP it is 512. All experimental results are averaged over five runs. It is also worth noting that FedDW requires removing the bias term parameters of the classification layer. This operation is only applied to FedDW, meaning that the classification layers of other frameworks remain as normal linear layers. We tune 
\(\mu\) from \{0.01, 0.1, 1, 10, 100\}.

\begin{figure}[h]
    \centering
    \begin{subfigure}[b]{0.35\textwidth}
        \centering
        \includegraphics[width=\textwidth]{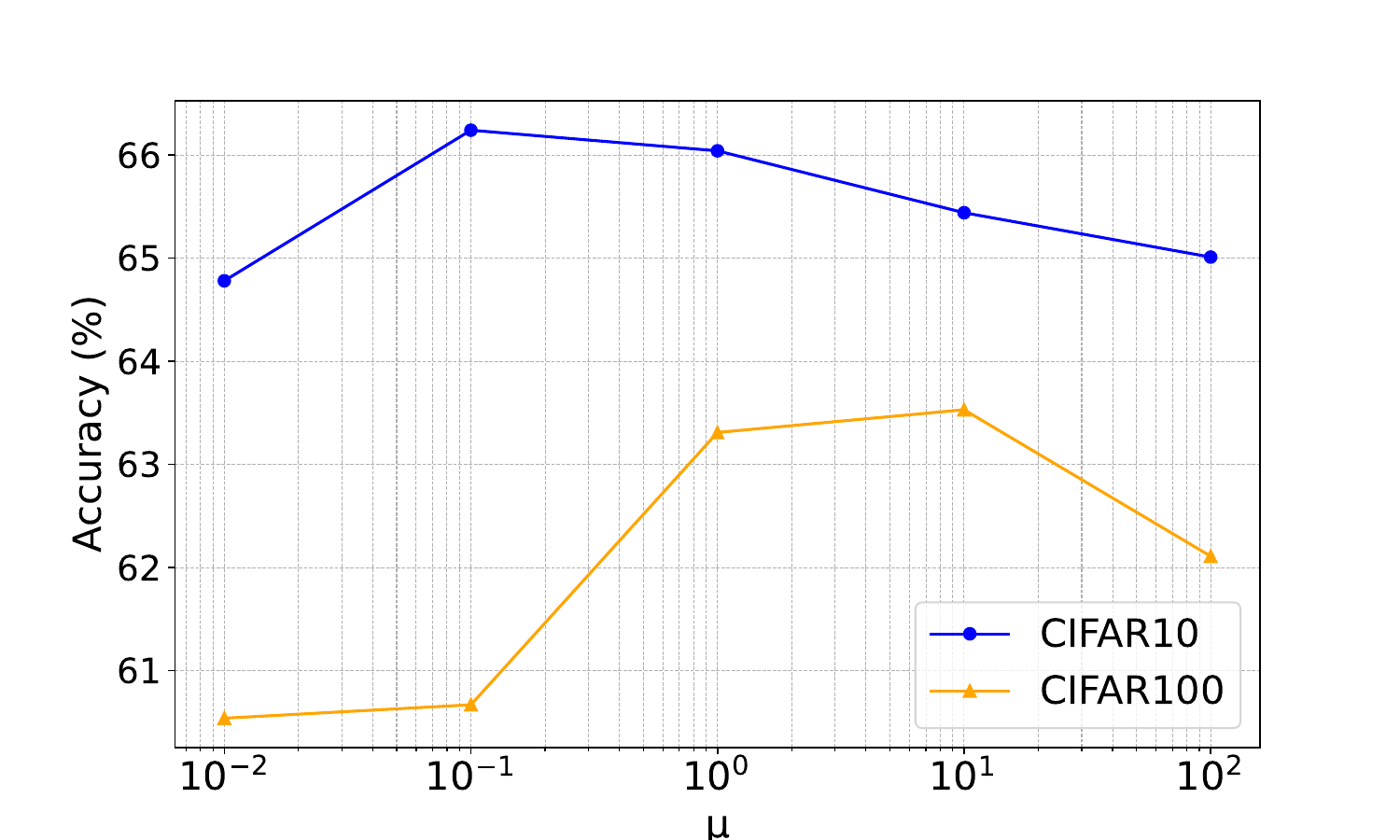}
        \caption{Practical Heterogeneous}
        \label{fig:second}
    \end{subfigure}\hfill
    \begin{subfigure}[b]{0.35\textwidth}
        \centering
        \includegraphics[width=\textwidth]{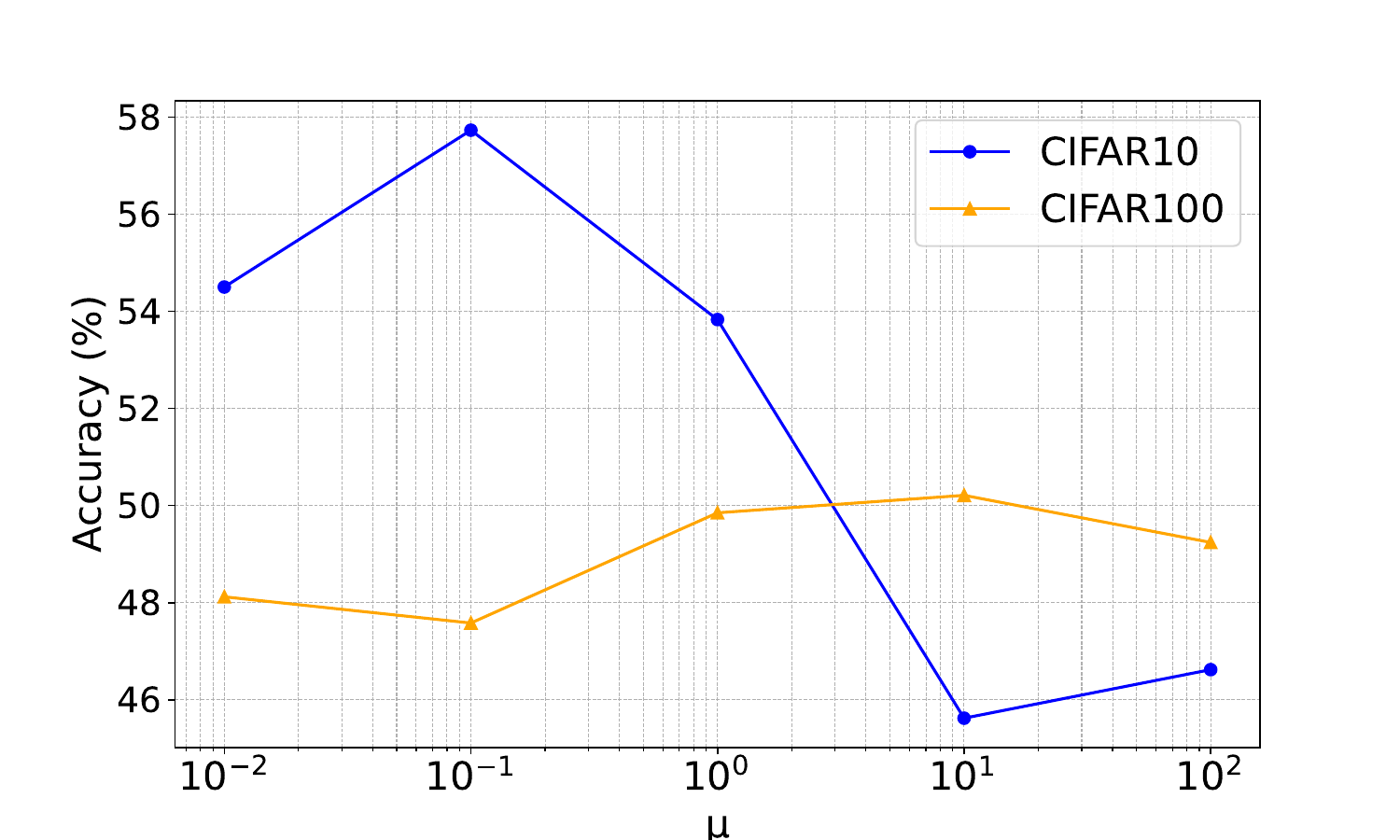}
        \caption{Pathological Heterogeneous}
        \label{fig:third}
    \end{subfigure}
    \caption{Accuracy for different \(\mu\) values on CIFAR10 and CIFAR100}
    \label{fig:6}
\end{figure}

To simulate heterogeneous data distribution, we use the popular Dirichlet distribution 
\(Dir(\beta)\) to form non-IID data distribution, where the parameter \(\beta\) controls the degree of data heterogeneity. The smaller the \(\beta\), the greater the degree of heterogeneity. 

\subsection{Accuracy Comparison}
We designed two different heterogeneous environments for comparison experiments: Practical Heterogeneous and Pathological Heterogeneous. In Practical Heterogeneous, the \(\beta\) is 0.5, and the user participation rate per communication round is 1. In the Pathological Heterogeneous environment, the \(\beta\) is 0.1, and the participation rate is 0.5. Additionally, in the CV domain, the number of communication rounds is set to 50, and the local training epoch for clients is 5. In the NLP domain, we set the number of communication rounds to 10 because we use the IMDB dataset for few-shot evaluation. A framework with high accuracy is not necessarily the best; in practical applications, a FL framework should achieve high accuracy with fewer communication rounds.
\begin{table*}[h!]
\caption{Comparison of different FL frameworks under practical and pathological heterogeneous settings.The font deepening represents the best result in the same environment, and superscript * represents the second best result and - means that the model crashes.}
\centering
\begin{tabular}{cccc|c|ccc|c}
\hline\hline
\multirow{3}{*}{\textbf{Method}} & \multicolumn{4}{c|}{\textbf{Practical Heterogeneous}} & \multicolumn{4}{c}{\textbf{Pathological Heterogeneous}} \\ \cline{2-9}
 & \multicolumn{3}{c|}{\textbf{CV}} & \textbf{NLP} & \multicolumn{3}{c|}{\textbf{CV}} & \textbf{NLP} \\ \cline{2-9}
 & \textbf{MNIST} & \textbf{CIFAR-10} & \textbf{CIFAR-100} & \textbf{IMDB} & \textbf{MNIST} & \textbf{CIFAR-10} & \textbf{CIFAR-100} & \textbf{IMDB} \\ \hline
FedAvg & 99.31$\pm$0.2 & 63.44$\pm$0.4 & 62.16$\pm$0.6 & 83.33$\pm$0.4* & 95.05$\pm$0.5 & 48.36$\pm$0.3 & 43.47$\pm$0.5 & 71.09$\pm$0.9 \\ 
MOON & 99.42$\pm$0.1 & 64.33$\pm$0.5 & 63.35$\pm$0.3 & 81.14$\pm$0.3 & 96.11$\pm$0.1 & 54.01$\pm$0.4 & 49.04$\pm$0.2 & 80.28$\pm$0.4 \\ 
FedBABU & 99.51$\pm$0.1* & 65.59$\pm$0.2* & 63.49$\pm$0.2* & 82.41$\pm$0.4 & \textbf{96.78$\pm$0.2} & 53.06$\pm$0.2 & 50.12$\pm$0.3* & 81.08$\pm$0.2* \\ 
FedUV & \textbf{99.52$\pm$0.2} & 65.05$\pm$0.3 & 63.35$\pm$0.2 & 68.22$\pm$0.6 & 96.65$\pm$0.3 & \textbf{54.97$\pm$0.3*} & 44.09$\pm$0.5 & 55.12$\pm$0.4 \\ 
FedProx & 99.39$\pm$0.1 & 64.74$\pm$0.2 & 62.81$\pm$1.0 & 73.12$\pm$0.2 & 95.12$\pm$0.4 & 51.39$\pm$0.3 & 46.07$\pm$0.2 & 66.84$\pm$0.4 \\ 
FedRep & 96.88$\pm$0.3 & 52.42$\pm$1.1 & 39.03$\pm$2.2 & 74.64$\pm$1.2 & 67.23$\pm$1.2 & 28.05$\pm$1.5 & 23.46$\pm$0.9 & 54.13$\pm$0.8 \\ 
FedProto & 97.11$\pm$0.1 & 45.52$\pm$0.8 & 23.15$\pm$0.9 & 70.69$\pm$0.8 & 49.83$\pm$4.2 & 27.64$\pm$1.6 & 16.48$\pm$1.3 & 53.85$\pm$2.4 \\ 
FedALA & 99.02$\pm$0.1 & 48.39$\pm$1.5 & 41.97$\pm$3.3 & 74.14$\pm$0.8 & 79.16$\pm$1.2 & 32.58$\pm$0.8 & 28.16$\pm$1.2 & 61.46$\pm$0.7 \\ 
PerFedAvg & 99.12$\pm$0.1 & 64.23$\pm$0.1 & 62.22$\pm$0.5 & 79.76$\pm$0.6 & 96.21$\pm$0.5 & - & - & 63.45$\pm$0.8 \\ \hline
FedPer & 96.55$\pm$0.5 & 48.78$\pm$1.5 & 36.87$\pm$1.2 & 50.87$\pm$0.6 & 64.89$\pm$2.2 & 30.38$\pm$1.3 & 22.82$\pm$4.2 & 53.08$\pm$1.4 \\ 
Local & 95.35$\pm$3.2 & 42.38$\pm$5.1 & 22.83$\pm$6.2 & 63.18$\pm$3.6 & 55.76$\pm$9.2 & 25.75$\pm$3.2 & 15.54$\pm$8.8 & 51.11$\pm$4.5 \\ 
FedDW & \textbf{99.52$\pm$0.1} & \textbf{66.24$\pm$0.6} &\textbf{63.53$\pm$0.3} & \textbf{84.46$\pm$0.3} & 96.66$\pm$0.5* & \textbf{57.73$\pm$0.5} & \textbf{50.21$\pm$0.3} & \textbf{82.81$\pm$0.4} \\ \hline\hline
\end{tabular}
\label{tab:1}
\end{table*}

\begin{table*}[ht]
    \centering
    \caption{Scalability test results. Baseline1 refers to CIFAR-10 experimental data under Practical Heterogeneous conditions from Table~\ref{tab:1}, while Baseline2 refers to CIFAR-100 data under the same conditions.}
    \begin{tabular}{lc|cc|cc|c|cccccc}
        \toprule
        \hline
        \multirow{2}{*}{Method} & \multicolumn{1}{c}{Baseline1} & \multicolumn{2}{c}{Client Quantity Scalability} & \multicolumn{2}{c}{Training Rounds Scalability} & \multicolumn{1}{c}{Baseline2} & \multicolumn{2}{c}{Models Scalability} \\
        \cmidrule(lr){2-2} \cmidrule(lr){3-4} \cmidrule(lr){5-6} \cmidrule(lr){7-7} \cmidrule(lr){8-9}
        & standard & N=50 & N=100 & Sr=25 Cr=10 & Sr=125 Cr=2 & ResNet18 & ShuffleNet & MobileNet \\
        \midrule
        FedAvg & 63.4$\pm$0.4 & 56.07$\pm$0.1 & 51.41$\pm$0.3* & 65.72$\pm$0.3 & 66.09$\pm$0.2* & 62.16$\pm$0.6 & 50.41$\pm$0.4 & 52.77$\pm$0.5 \\
        MOON & 64.33$\pm$0.5 & 53.59$\pm$0.4 & 49.59$\pm$0.5 & 65.84$\pm$0.4 & 65.76$\pm$0.7 & 63.35$\pm$0.3 & 50.87$\pm$0.3 & 52.98$\pm$0.5 \\
        FedBABU & 65.59$\pm$0.2* & 56.12$\pm$0.2 & 50.55$\pm$0.1 & 66.89$\pm$0.1* & 66.33$\pm$0.2* & 63.49$\pm$0.2* & \textbf{52.08$\pm$0.2} & 53.05$\pm$0.4 \\
        FedUV & 65.05$\pm$0.3 & 56.81$\pm$0.2* & - & 66.81$\pm$0.3 & 67.01$\pm$0.4* & 63.35$\pm$0.2 & 51.97$\pm$0.2 & 53.13$\pm$0.1* \\
        FedProx & 64.74$\pm$0.2 & 55.87$\pm$0.6 & 51.19$\pm$0.5 & 65.84$\pm$0.2 & 66.19$\pm$0.4 & 62.81$\pm$0.6 & 50.34$\pm$0.3 & 50.52$\pm$0.2 \\
        FedRep & 52.42$\pm$1.1 & 38.66$\pm$1.3 & 34.68$\pm$1.4 & 45.84$\pm$0.3 & 45.19$\pm$0.4 & 39.03$\pm$1.1 & 29.17$\pm$0.6 & 41.66$\pm$0.7 \\
        FedProto & 45.52$\pm$0.8 & 37.15$\pm$1.1 & 34.57$\pm$0.8 & 44.52$\pm$0.6 & 45.04$\pm$0.3 & 23.15$\pm$0.9 & 28.34$\pm$1.4 & 30.43$\pm$0.8 \\
        FedALA & 48.39$\pm$1.5 & 39.71$\pm$1.0 & 35.87$\pm$0.7 & 46.55$\pm$0.3 & 47.54$\pm$0.2 & 41.97$\pm$0.5 & 34.17$\pm$0.4 & 32.64$\pm$0.6 \\
        PerFedAvg & 64.23$\pm$0.1 & 55.23$\pm$0.3 & 51.10$\pm$0.3 & 63.55$\pm$0.2 & 64.53$\pm$0.5 & 62.22$\pm$0.5 & 48.44$\pm$0.3 & 49.11$\pm$0.3 \\
        FedPer & 48.78$\pm$1.5 & 40.84$\pm$2.2 & 36.78$\pm$1.9 & 46.68$\pm$1.5 & 47.74$\pm$1.9 & 36.87$\pm$1.2 & 35.22$\pm$1.4 & 35.89$\pm$0.9 \\
        Local & 42.38$\pm$1.5 & 28.13$\pm$1.2 & 28.13$\pm$1.2 & 40.68$\pm$3.4 & 41.02$\pm$2.8 & 22.83$\pm$6.2 & 25.71$\pm$1.2 & 27.94$\pm$0.5 \\
        FedDW & \textbf{66.24$\pm$0.6} & \textbf{57.6$\pm$0.3} & \textbf{51.73$\pm$0.4} & \textbf{67.06$\pm$0.4} & \textbf{67.21$\pm$0.2} & \textbf{63.53$\pm$0.3} & 51.92$\pm$0.4* & \textbf{53.71$\pm$0.3} \\\hline
        \bottomrule
    \end{tabular}
    \label{tab:2}
\end{table*}

\begin{table*}
\centering
\caption{The time it takes to run one epoch on the client side using the ResNet18 model on the  CIFAR-100 dataset in Pathological Heterogeneous environment by one RTX4060 GPU.}
\begin{tabular}{lcccccccc}
\hline
\textbf{Method} & \textbf{FedAvg} & \textbf{MOON} & \textbf{FedBABU} & \textbf{FedUV} & \textbf{PerFedAvg} & \textbf{FedProx} & \textbf{FedDW}\\
\hline
Time (ms/epoch) & 16.05	& 18.93	& 15.67	 & 17.18 & 19.17 & 17.32 & 16.11\\
\hline
\end{tabular}
\label{tab:3}
\end{table*}

\begin{figure*}
\centering
\includegraphics[width=\textwidth]{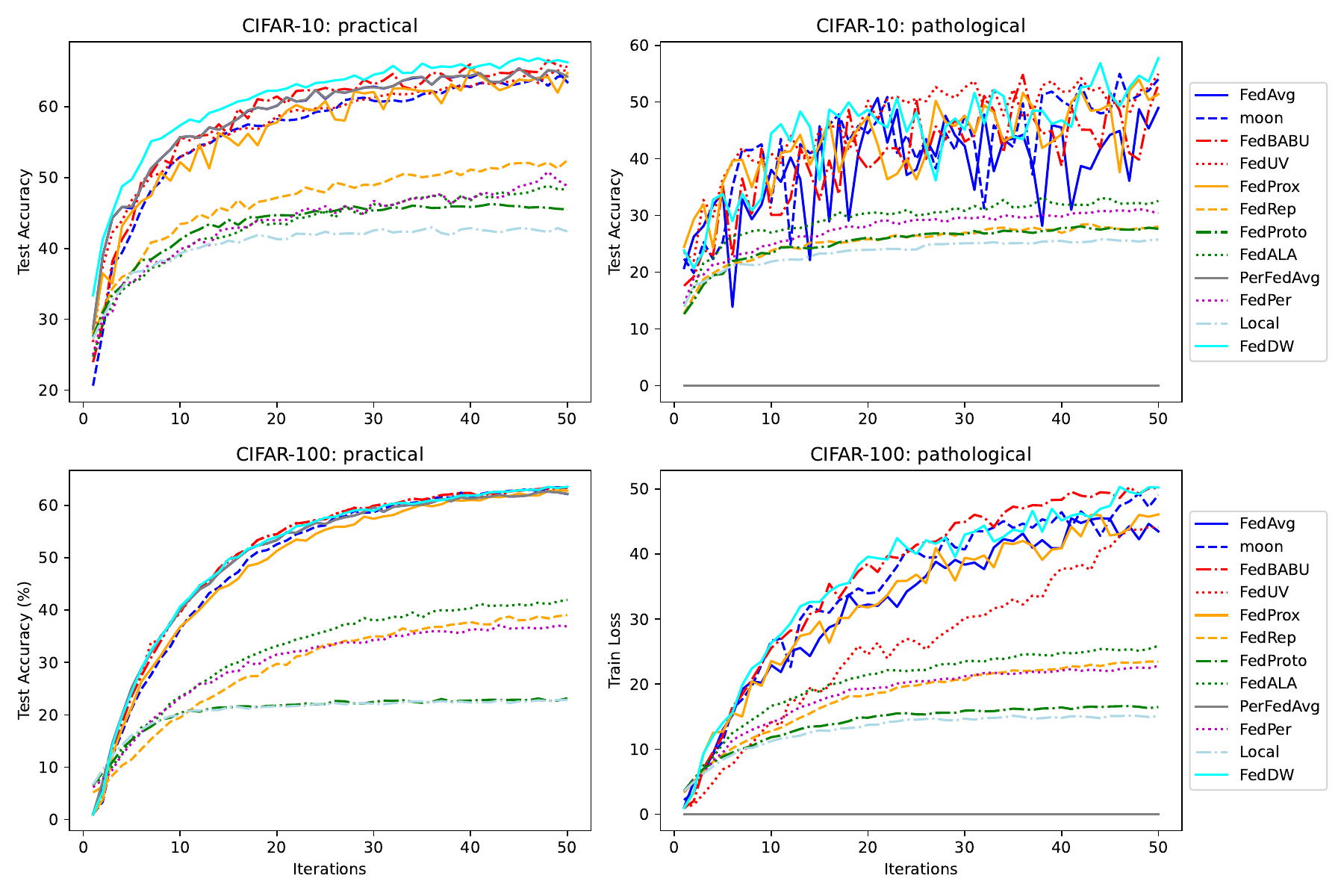}
\caption{Test accuracy under different communication rounds in both Pathological Heterogeneous and Practical Heterogeneous environments on the CIFAR-10 and CIFAR-100 datasets.}
\label{fig:7}
\end{figure*}

The experimental results are shown in Table~\ref{tab:1} and Figure~\ref{fig:6}, with training curves for some results illustrated in Figure~\ref{fig:7}. Firstly, the optimal value of $\mu$ varies across datasets, with $\mu = 0.1$ performing best on CIFAR-10 and $\mu = 10$ on CIFAR-100. The other two datasets have not changed much. Second, other experimental results demonstrate that FedDW consistently delivers top-tier or highly competitive performance under all experimental conditions. Additionally, as data heterogeneity increases, the accuracy of all methods decreases. However, the decline in our method is the smallest, with only 8.51\% and 13.32\% drops on the CIFAR-10 and CIFAR-100 datasets, respectively.
In terms of few-shot evaluation, FedDW also achieved the best results. Combined with the learning curves in Figure~\ref{fig:7}, it can be proved that our method has faster convergence. We also note that personalized FL methods often have low accuracy in global tests conducted on the server because they typically do not share their backbone network with the server to reduce communication bandwidth. This creates a trade-off between communication bandwidth and accuracy.

\subsection{Scalability}
To better demonstrate the scalability of the FedDW framework, we conducted Client Quantity Scalability experiments and Training Rounds Scalability experiments using the CIFAR-10 dataset in the Practical Heterogeneous environment, with the CIFAR-10 experiment data serving as Baseline1. Additionally, we performed Models Scalability experiments using the CIFAR-100 dataset (with ResNet-18 as the model) as Baseline2. The experimental results are shown in Table~\ref{tab:2}.
\subsubsection{Client Quantity Scalability}
In the Client Quantity Scalability experiment, we varied the number of clients participating in the training. In Baseline1, the number of clients is 10. We increased this number to 50 and 100. We observed that as the number of clients increases, the accuracy correspondingly decreases. This is expected because the total amount of data remains the same, so each client receives less data, resulting in poorer model generalization. Under both client numbers, our method still achieves the highest accuracy.
\subsubsection{Training Rounds Scalability}
In Baseline1, the local training epochs Cr are 5, and the communication rounds Sr are 50, making the total number of epochs for all clients 250 throughout the experiment. We maintained this product while dynamically changing the values of Cr and Sr. We selected two sets of values: (Sr=25, Cr=10) and (Sr=125, Cr=2), representing two extreme cases. The data in Table~\ref{tab:2} shows that our method achieved the best results in both cases, with only a 0.14\% difference in accuracy between the two extreme scenarios. Additionally, we observed that the highest accuracy achieved in the second case is 67.2\%.
\subsubsection{Models Scalability}
In the Models Scalability experiment, we applied two different neural network models, ShuffleNet\cite{b45} and MobileNet\cite{b46}, on the CIFAR-100 dataset, and compared them with the ResNet-18 model from Baseline1. Our method achieved the best performance with MobileNet, while it achieved the second-best performance with ShuffleNet, with only a 0.16\% difference from the best-performing method, FedBABU. Combined with the results of using LSTM on the IMDB dataset from Table~\ref{tab:1}, it can be demonstrated that our method has excellent model scalability.
\begin{figure*}[h]
    \centering
    \begin{subfigure}[b]{0.33\textwidth}
        \centering
        \includegraphics[width=\textwidth]{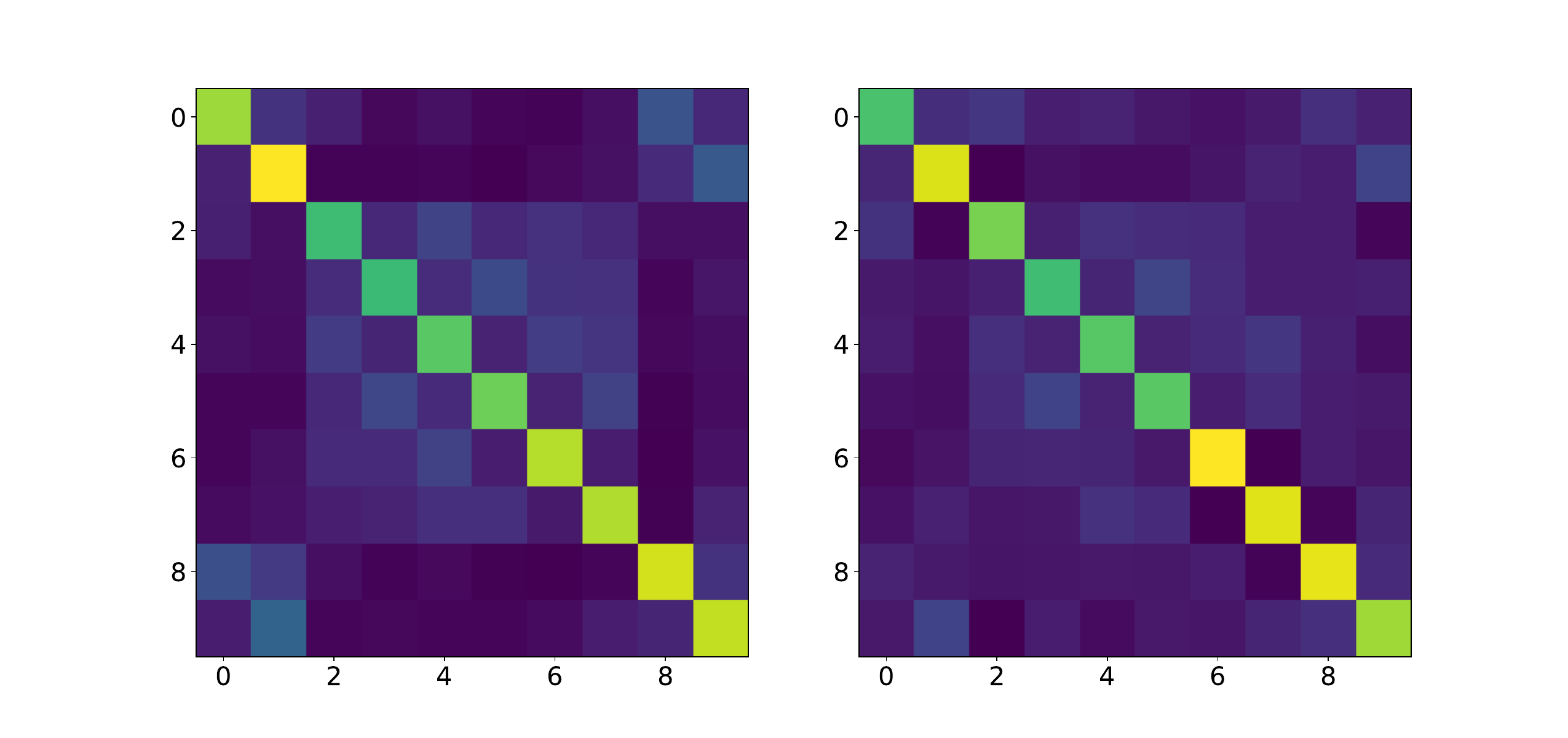}
        \caption{Fedavg IID}
        \label{fig:first}
    \end{subfigure}\hfill
    \begin{subfigure}[b]{0.33\textwidth}
        \centering
        \includegraphics[width=\textwidth]{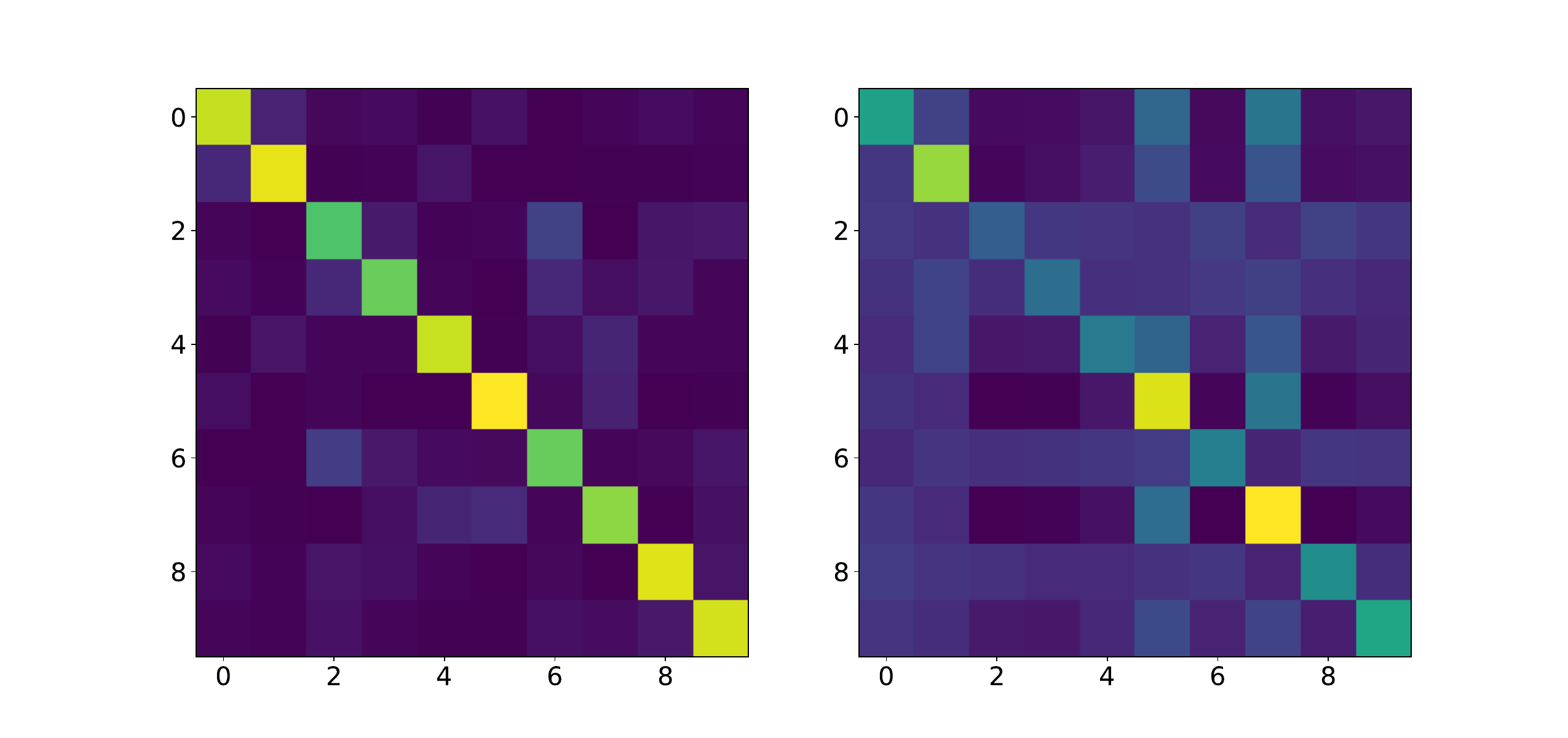}
        \caption{Fedavg non-IID}
        \label{fig:second}
    \end{subfigure}\hfill
    \begin{subfigure}[b]{0.33\textwidth}
        \centering
        \includegraphics[width=\textwidth]{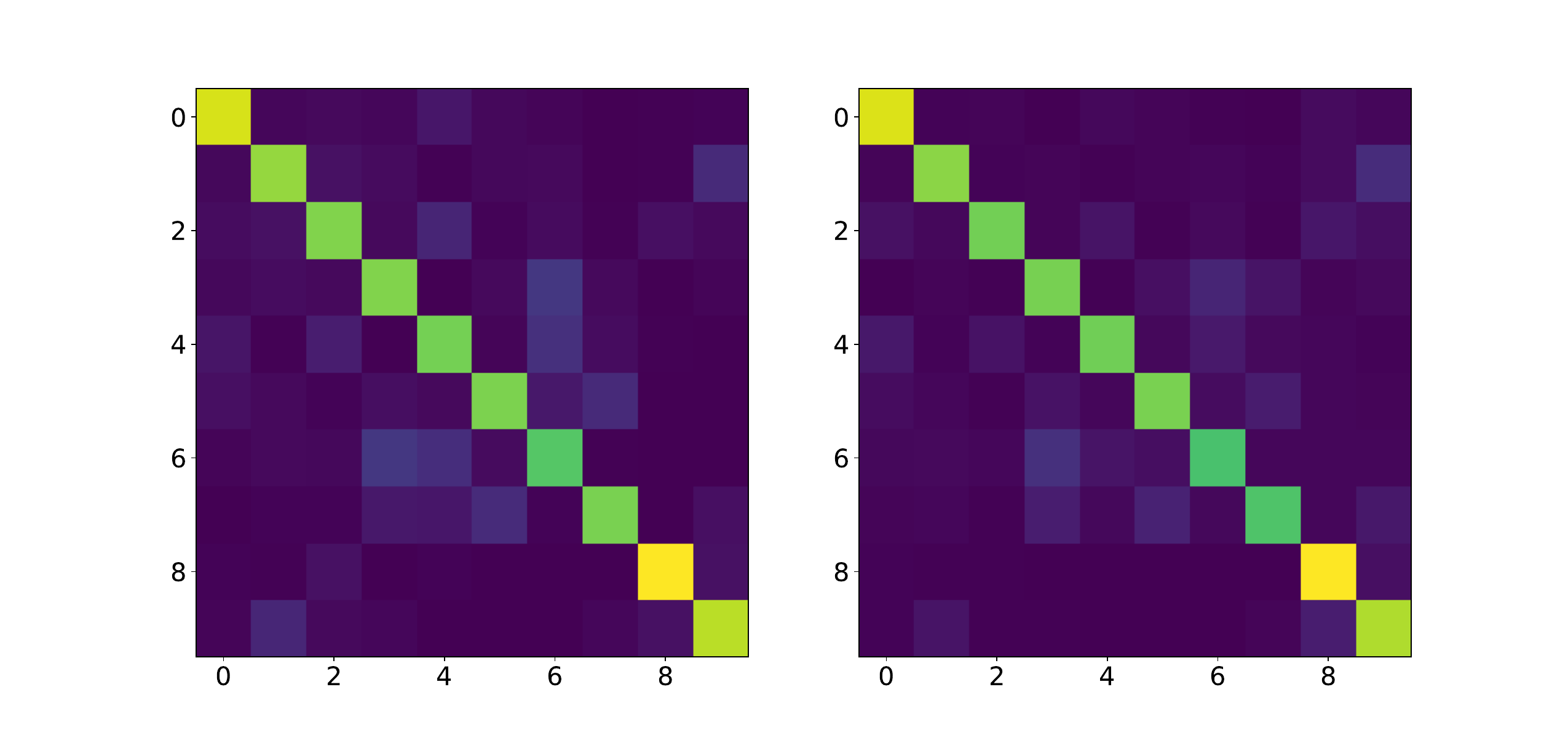}
        \caption{FedDW non-IID}
        \label{fig:third}
    \end{subfigure}
    \caption{Heatmap Visualization of SL matrix(Left) and CR matrix(Right) under three different conditions. Non-IID refers to the pathological heterogeneous environment mentioned earlier.}
    \label{fig:8}
\end{figure*}

\subsection{Efficiency}
The efficiency of FL frameworks consists of two main parts: communication efficiency and local computation efficiency. In FedDW, based on FedAvg, clients additionally share the SL matrix and the number of data for each class with the server. The additional communication overhead is \( 2(|\mathcal{C}|+|\mathcal{C}|^{2})\), which is negligible compared to the communication overhead of the entire neural network model. As in the above analysis, the regularization term added by our method to mitigate data heterogeneity only involves the classifier parameters. Compared to other non-pFL methods, this property reduces the computational burden of backpropagation during local training.

In Table~\ref{tab:3}, we summarize the time taken for one round of local training using the ResNet-18 model on the CIFAR-100 dataset for seven non-personalized FL methods. Our method, apart from FedBABU, is the closest to Fedavg. The difference in training time between each round is only 0.06 milliseconds.

\subsection{Compatibility}
The regularization term of FedDW only involves the network parameters of the classification layer, so the optimization method of FedDW can be better combined with other FL optimization methods. To a large extent, this avoids the inherent conflicts of different optimization methods. We abstract the optimization method of FedDW as DW, combine it with other FL methods, and experiment with its accuracy.


\begin{table}[h]
    \centering
     \caption{Comparison of methods and mixed methods with ACC results under pathological heterogeneous settings in CIFAR-100 dataset. Fedistill+ adds the entire network global aggregation on the basis of \cite{b17}. 
     }
    \resizebox{\columnwidth}{!}{
    \begin{tabular}{|c|c|c|c|c|c|}
        \hline
        \textbf{Method} & \textbf{Fedavg} & \textbf{FedDistill+} & \textbf{MOON} & \textbf{FedUV} & \textbf{Fedprox} \\ \hline
        \textbf{ACC} & 43.47±0.5 & 43.96±0.8 & 49.04±0.9 & 44.09±0.5 & 46.07±0.3 \\ \hline
        \textbf{Mixed Method} & FedDW & FedDistill++DW & MOON+DW & FedUV+DW & Fedprox+DW \\ \hline
        \textbf{ACC} & \textbf{50.21±0.3} & \textbf{47.26±0.6} & \textbf{52.81±0.2} & \textbf{44.09±0.8} & \textbf{48.39±0.8} \\ \hline
    \end{tabular}
    }
     \label{tab:4}
\end{table}

The experimental results are shown in Table~\ref{tab:4} . 
The DW optimization module generally improves the results in most cases. Notably, combining it with FedUV showed no change in experimental outcomes. We speculate that this is because the essence of FedUV optimization is to refine neural network parameter structures, ensuring that they maintain IID distribution even on heterogeneous data. This aligns with our optimization approach, differing only in methodology. As both methods effectively achieve the same objective, the accuracy remains unchanged.

\subsection{The CR matrix and The SL matrix}
We visualized the SL and CR matrices under three different conditions using heatmaps to explore their properties. The visualization results are shown in Figure~\ref{fig:8}. Under IID conditions, the SL and CR matrices obtained by FedAvg have high similarity. However, under non-IID conditions, there is no similarity between the two. Nonetheless, we can also observe that the SL matrices under both conditions are almost similar. This supports our idea that the SL matrix maintains global generalization under non-IID conditions and regularizes the CR matrix using the SL matrix reasonably well. In non-IID environments, FedDW’s SL and CR matrices both resemble those under IID conditions, and the two matrices are similar to each other as well. This demonstrates that our method can mitigate the issue of data heterogeneity.

\section{Conclusion}
FedDW is a simple and effective FL method. It effectively mitigates the challenges posed by data heterogeneity with minimal additional computational and communication costs. Moreover, extensive scalability experiments have demonstrated the robustness of our approach. Additionally, because the regularization cost in FedDW does not increase with the model size, our method may be particularly well-suited for the current era of large models. However, due to experimental constraints, we have not yet tested or analyzed the application of FedDW in large-scale FL scenarios. Furthermore, since many deep learning tasks do not require a classification layer, the applicability of FedDW is somewhat limited. In future work, we plan to apply the FedDW framework to large-scale model training and extend its application to a broader range of deep learning tasks.

\section*{Acknowledgment}
This work is supported by the Humanities and Social Sciences Youth Foundation, Ministry of Education of China (Grant No.20YJCZH172), the China Postdoctoral Science Foundation (Grant No.2019M651262), the Heilongjiang Provincial Postdoctoral Science Foundation (Grant No.LBHZ19015), and funding from the Basic Research Support Program for Outstanding Young Teachers in Provincial Undergraduate Universities in Heilongjiang Province (YQJH2023302).



\begin{thebibliography}{00}
\bibitem{b1}K. Zhang, X. Song, C. Zhang, and S. Yu, ``Challenges and future directions of secure federated learning: A survey," \textit{Front. Comput. Sci.}, vol. 16, pp. 1--8, 2022.

\bibitem{b2}P. Kairouz, H. B. McMahan, B. Avent, A. Bellet, M. Bennis, A. N. Bhagoji, K. Bonawitz, Z. Charles, G. Cormode, R. Cummings, \textit{et al.}, ``Advances and open problems in federated learning," \textit{Found. Trends Mach. Learn.}, vol. 14, no. 1--2, pp. 1--210, 2021.

\bibitem{b3}M. Ye, X. Fang, B. Du, P. C. Yuen, and D. Tao, ``Heterogeneous federated learning: State-of-the-art and research challenges," \textit{ACM Comput. Surveys}, vol. 56, no. 3, pp. 1--44, 2023.

\bibitem{b4}B. McMahan, E. Moore, D. Ramage, S. Hampson, and B. Aguera y Arcas, ``Communication-efficient learning of deep networks from decentralized data," in \textit{Proc. Artif. Intell. Statist.}, 2017, pp. 1273--1282.

\bibitem{b5}K. Guo, Y. Ding, J. Liang, R. He, Z. Wang, and T. Tan, ``Not all minorities are equal: Empty-class-aware distillation for heterogeneous federated learning," \textit{arXiv preprint arXiv:2401.02329}, 2024.

\bibitem{b6}T. Li, A. K. Sahu, M. Zaheer, M. Sanjabi, A. Talwalkar, and V. Smith, ``Federated optimization in heterogeneous networks," \textit{Proc. Mach. Learn. Syst.}, vol. 2, pp. 429--450, 2020.

\bibitem{b7}M. Ye, X. Fang, B. Du, P. C. Yuen, and D. Tao, ``Heterogeneous federated learning: State-of-the-art and research challenges," \textit{ACM Comput. Surveys}, vol. 56, no. 3, pp. 1--44, 2023.

\bibitem{b8}Y. Zhao, M. Li, L. Lai, N. Suda, D. Civin, and V. Chandra, ``Federated learning with non-IID data," \textit{arXiv preprint arXiv:1806.00582}, 2018.

\bibitem{b9}J. Shi, S. Zheng, X. Yin, Y. Lu, Y. Xie, and Y. Qu, ``Clip-guided federated learning on heterogeneous and long-tailed data," \textit{arXiv preprint arXiv:2312.08648}, 2023.

\bibitem{b10}B. Kang, S. Xie, M. Rohrbach, Z. Yan, A. Gordo, J. Feng, and Y. Kalantidis, ``Decoupling representation and classifier for long-tailed recognition," \textit{arXiv preprint arXiv:1910.09217}, 2019.

\bibitem{b11}L. Yi, G. Wang, X. Liu, Z. Shi, and H. Yu, ``FedGH: Heterogeneous federated learning with generalized global header," in \textit{Proc. 31st ACM Int. Conf. Multimedia}, 2023, pp. 8686--8696.

\bibitem{b12}J. Zhang, Y. Liu, Y. Hua, and J. Cao, ``FedTGP: Trainable Global Prototypes with Adaptive-Margin-Enhanced Contrastive Learning for Data and Model Heterogeneity in Federated Learning," in \textit{Proceedings of the AAAI Conference on Artificial Intelligence}, vol. 38, no. 15, pp. 16768-16776, 2024.


\bibitem{b13}S. Han, S. Park, F. Wu, S. Kim, C. Wu, X. Xie, and M. Cha, ``Fedx: Unsupervised federated learning with cross knowledge distillation," in \textit{Proc. Eur. Conf. Comput. Vis. (ECCV)}, 2022, pp. 691--707.

\bibitem{b14}Y. Tan, G. Long, L. Liu, T. Zhou, Q. Lu, J. Jiang, and C. Zhang, ``Fedproto: Federated prototype learning across heterogeneous clients," in \textit{Proc. AAAI Conf. Artif. Intell.}, vol. 36, no. 8, 2022, pp. 8432--8440.


\bibitem{b15}C. Xie, D.-A. Huang, W. Chu, D. Xu, C. Xiao, B. Li, and A. Anandkumar, ``PerAda: Parameter-Efficient Federated Learning Personalization with Generalization Guarantees," in \textit{Proc. IEEE/CVF Conf. Comput. Vis. Pattern Recognit. (CVPR)}, 2024, pp. 23838–23848.

\bibitem{b16}L. Xie, M. Lin, T. Luan, C. Li, Y. Fang, Q. Shen, and Z. Wu, ``MH-pFLID: Model heterogeneous personalized federated learning via injection and distillation for medical data analysis," \textit{arXiv preprint arXiv:2405.06822}, 2024.

\bibitem{b17}E. Jeong, S. Oh, H. Kim, J. Park, M. Bennis, and S.-L. Kim, ``Communication-efficient on-device machine learning: Federated distillation and augmentation under non-IID private data," \textit{arXiv preprint arXiv:1811.11479}, 2018.


\bibitem{b18}G. Hinton, ``Distilling the knowledge in a neural network," \textit{arXiv preprint arXiv:1503.02531}, 2015.


\bibitem{b19}A. E. Durmus, Z. Yue, M. Ramon, M. Mattina, P. Whatmough, and V. Saligrama, ``Federated learning based on dynamic regularization," in \textit{Proc. Int. Conf. Learn. Representations (ICLR)}, 2021.


\bibitem{b20}Q. Li, B. He, and D. Song, ``Model-contrastive federated learning," in \textit{Proc. IEEE/CVF Conf. Comput. Vis. Pattern Recognit.}, 2021, pp. 10713--10722.

\bibitem{b21}L. Zhu, Z. Liu, and S. Han, ``Deep leakage from gradients," \textit{Adv. Neural Inf. Process. Syst.}, vol. 32, 2019.

\bibitem{b22}K. He, H. Fan, Y. Wu, S. Xie, and R. Girshick, ``Momentum contrast for unsupervised visual representation learning," in \textit{Proc. IEEE/CVF Conf. Comput. Vis. Pattern Recognit.}, 2020, pp. 9729--9738.

\bibitem{b23}Y. Yang, C. Huang, L. Xia, C. Huang, D. Luo, and K. Lin, ``Debiased contrastive learning for sequential recommendation," in \textit{Proc. ACM Web Conf. 2023}, 2023, pp. 1063--1073.

\bibitem{b24}Y. Zheng, X. Zhao, C. Lan, X. Zhang, B. Huang, J. Yang, and D. Yu, ``CPCL: Cross-modal prototypical contrastive learning for weakly supervised text-based person re-identification," \textit{arXiv preprint arXiv:2401.10011}, 2024.

\bibitem{b25}T. Chen, S. Kornblith, M. Norouzi, and G. Hinton, ``A simple framework for contrastive learning of visual representations," in \textit{Proc. Int. Conf. Mach. Learn. (ICML)}, 2020, pp. 1597--1607.

\bibitem{b26} J. Zhang, Y. Liu, Y. Hua, and J. Cao, ``An Upload-Efficient Scheme for Transferring Knowledge From a Server-Side Pre-trained Generator to Clients in Heterogeneous Federated Learning," in \textit{Proc. IEEE/CVF Conf. Comput. Vis. Pattern Recognit. (CVPR)}, 2024, pp. 12109–12119.

\bibitem{b27}Y. Chen, W. Huang, and M. Ye, ``Fair federated learning under domain skew with local consistency and domain diversity," in \textit{Proc. IEEE/CVF Conf. Comput. Vis. Pattern Recognit.}, 2024, pp. 12077--12086.

\bibitem{b28}H. Wang, Y. Li, W. Xu, R. Li, Y. Zhan, and Z. Zeng, ``Dafkd: Domain-aware federated knowledge distillation," in \textit{Proc. IEEE/CVF Conf. Comput. Vis. Pattern Recognit.}, 2023, pp. 20412–20421.

\bibitem{b29}F. Sattler, T. Korjakow, R. Rischke, and W. Samek, ``Fedaux: Leveraging unlabeled auxiliary data in federated learning," \textit{IEEE Trans. Neural Netw. Learn. Syst.}, vol. 34, no. 9, pp. 5531--5543, 2021.

\bibitem{b30}L. Collins, H. Hassani, A. Mokhtari, and S. Shakkottai, ``Exploiting shared representations for personalized federated learning," in \textit{Proc. Int. Conf. Mach. Learn. (ICML)}, 2021, pp. 2089--2099.

\bibitem{b31}J. Zhang, Y. Hua, H. Wang, T. Song, Z. Xue, R. Ma, and H. Guan, ``Fedala: Adaptive local aggregation for personalized federated learning," in \textit{Proc. AAAI Conf. Artif. Intell.}, vol. 37, no. 9, 2023, pp. 11237--11244.

\bibitem{b32}M. G. Arivazhagan, V. Aggarwal, A. K. Singh, and S. Choudhary, ``Federated learning with personalization layers," \textit{arXiv preprint arXiv:1912.00818}, 2019.

\bibitem{b33}Z. Li, G. Long, and T. Zhou, ``Federated recommendation with additive personalization," in \textit{Proc. 12th Int. Conf. Learn. Representations (ICLR)}, 2024. [Online]. Available: https://openreview.net/forum?id=xkXdE81mOK.

\bibitem{b34}A. Z. Tan, H. Yu, L. Cui, and Q. Yang, ``Towards personalized federated learning," \textit{IEEE Trans. Neural Netw. Learn. Syst.}, vol. 34, no. 12, pp. 9587--9603, 2022.

\bibitem{b35}S. Minaee, T. Mikolov, N. Nikzad, M. Chenaghlu, R. Socher, X. Amatriain, and J. Gao, ``Large language models: A survey," \textit{arXiv preprint arXiv:2402.06196}, 2024.


\bibitem{b36} L. Wu, Z. Zheng, Z. Qiu, H. Wang, H. Gu, T. Shen, C. Qin, C. Zhu, H. Zhu, Q. Liu, \textit{et al.}, ``A survey on large language models for recommendation," \textit{World Wide Web}, vol. 27, no. 5, p. 60, 2024.

\bibitem{b37}Y. Chang, X. Wang, J. Wang, Y. Wu, L. Yang, K. Zhu, H. Chen, X. Yi, C. Wang, Y. Wang, \textit{et al.}, ``A survey on evaluation of large language models," \textit{ACM Trans. Intell. Syst. Technol.}, vol. 15, no. 3, pp. 1--45, 2024.

\bibitem{b38}M. Abadi, A. Chu, I. Goodfellow, H. B. McMahan, I. Mironov, K. Talwar, and L. Zhang, ``Deep learning with differential privacy," in \textit{Proc. ACM SIGSAC Conf. Comput. Commun. Secur.}, 2016, pp. 308--318.


\bibitem{b39}H. M. Son, M.-H. Kim, T.-M. Chung, C. Huang, and X. Liu, ``FedUV: Uniformity and variance for heterogeneous federated learning," in \textit{Proc. IEEE/CVF Conf. Comput. Vis. Pattern Recognit.}, 2024, pp. 5863–5872.

\bibitem{b40}J. Oh, S. Kim, and S.-Y. Yun, ``Fedbabu: Towards enhanced representation for federated image classification," \textit{arXiv preprint arXiv:2106.06042}, 2021.

\bibitem{b41}A. Fallah, A. Mokhtari, and A. Ozdaglar, ``Personalized federated learning: A meta-learning approach," \textit{arXiv preprint arXiv:2002.07948}, 2020.

\bibitem{b42}K. He, X. Zhang, S. Ren, and J. Sun, ``Deep residual learning for image recognition," in \textit{Proc. IEEE Conf. Comput. Vis. Pattern Recognit.}, 2016, pp. 770--778.


\bibitem{b43}A. Paszke, S. Gross, F. Massa, A. Lerer, J. Bradbury, G. Chanan, T. Killeen, Z. Lin, N. Gimelshein, L. Antiga, \textit{et al.}, ``Pytorch: An imperative style, high-performance deep learning library," \textit{Adv. Neural Inf. Process. Syst.}, vol. 32, 2019.

\bibitem{b44}D. P. Kingma and J. Ba, ``Adam: A Method for Stochastic Optimization," \textit{International Conference on Learning Representations (ICLR)}, San Diego, CA, USA, 2015. [Online]. Available: https://arxiv.org/abs/1412.6980


\bibitem{b45}X. Zhang, X. Zhou, M. Lin, and J. Sun, ``Shufflenet: An extremely efficient convolutional neural network for mobile devices," in \textit{Proc. IEEE Conf. Comput. Vis. Pattern Recognit.}, 2018, pp. 6848--6856.

\bibitem{b46}A. G. Howard, ``Mobilenets: Efficient convolutional neural networks for mobile vision applications," \textit{arXiv preprint arXiv:1704.04861}, 2017.

\end{thebibliography}
\end{document}